\documentclass[letterpaper]{article} 
\usepackage{aaai25}  
\usepackage{times}  
\usepackage{helvet}  
\usepackage{courier}  
\usepackage[hyphens]{url}  
\usepackage{graphicx} 
\usepackage{natbib}  
\usepackage{caption} 
\usepackage{algorithm}
\usepackage[noend]{algorithmic}
\usepackage{float}

\usepackage[utf8]{inputenc} 
\usepackage[T1]{fontenc}    
\usepackage{url}            
\usepackage{booktabs}       
\usepackage{amsfonts}       
\usepackage{nicefrac}       
\usepackage{microtype}      
\usepackage{graphicx}
\usepackage{xcolor}
\usepackage{natbib}
\usepackage{amsmath}
\usepackage{dblfloatfix}
\usepackage{newfloat}
\usepackage{listings}
\DeclareCaptionStyle{ruled}{labelfont=normalfont,labelsep=colon,strut=off} 
\floatstyle{ruled}
\newfloat{listing}{tb}{lst}{}
\floatname{listing}{Listing}
\title{TinySubNets: An Efficient and Low Capacity Continual Learning Strategy}
\author{
    Marcin Pietro\'n\textsuperscript{\rm 1}, 
    Kamil Faber\textsuperscript{\rm 1},
    Dominik Żurek\textsuperscript{\rm 1},
    Roberto Corizzo\textsuperscript{\rm 2}
}
\affiliations{
    \textsuperscript{\rm 1}AGH University of Krakow, Krakow, Poland\\
    \textsuperscript{\rm 2}American University, Washington DC, USA\\

    pietron@agh.edu.pl, kfaber@agh.edu.pl, 
    dzurek@agh.edu.pl,
    rcorizzo@american.edu
}

\usepackage{bibentry}

\begin{document}

\maketitle

\begin{abstract}
Continual Learning (CL) is a highly relevant setting gaining traction in recent machine learning research. Among CL works, architectural and hybrid strategies are particularly effective due to their potential to adapt the model architecture as new tasks are presented.
However, many existing solutions do not efficiently exploit model sparsity, and are prone to capacity saturation due to their inefficient use of available weights, which limits the number of learnable tasks. 
In this paper, we propose TinySubNets (TSN), a novel architectural CL strategy that addresses the issues through the unique combination of   pruning with different sparsity levels, adaptive quantization, and weight sharing. 
%
Pruning identifies a subset of weights that preserve model performance, making less relevant weights available for future tasks.
Adaptive quantization allows a single weight to be separated into multiple parts which can be assigned to different tasks.
Weight sharing between tasks boosts the exploitation of capacity and task similarity, allowing for the identification of a better trade-off between model accuracy and capacity.
These features allow TSN to efficiently leverage the available capacity, enhance knowledge transfer, and reduce   computational resources consumption. 
Experimental results involving common benchmark CL datasets and scenarios show that our proposed strategy achieves better results in terms of accuracy than existing state-of-the-art CL strategies. Moreover, our strategy is shown to provide a significantly improved model capacity exploitation. 
\end{abstract}

%
\begin{links}
\link{Code}{https://github.com/lifelonglab/tinysubnets}
\end{links}

%

\section{Introduction}
Continual learning (CL) is a machine learning paradigm aiming at designing effective methods and strategies to analyze data in complex and dynamic real-world environments \cite{parisi2019continual}. As models are challenged with multiple tasks over their lifetime, 
a desired property for CL strategies is to maintain a high performance across all tasks. This paradigm is receiving increasing attention, which led to many works being proposed \cite{parisi2019continual,BAKER2023274,FABER2023248,wortsman2020,zhou2023deep}. 
There are three main types of CL strategies: rehearsal (also known as experience replay), regularization, and architectural. 


Among architectural strategies, forget-free approaches are particularly relevant. They typically accommodate new tasks by assigning a subset of available model weights to each task. 
State-of-the-art approaches in this category include Packnet \cite{mallya2017}, WSN \cite{kang2022}, Ada-QPacknet \cite{pietron2023ada}, and DyTox \cite{douillard2022dytox}.
%
However, their effectiveness strongly depends on the degree to which they can exploit model capacity to accommodate as many tasks as possible. 
One important pitfall of most existing pruning-based architectural forget-free methods is that they are limited in their ability to exploit model sparsity, since each layer is pruned with the same constant sparsity level.
As a result, they do not properly tune the number of weights to be removed while preserving classification accuracy \cite{xu2021}.
A possible approach to address this issue is incorporating different sparsity levels for each layer \cite{pietron2023ada}. 
Another general limitation of recent forget-free methods is that they are prone to a quick saturation of the available model capacity in that they cannot assign more than one value to each weight and are, therefore, limited by the available weights, as in \cite{douillard2022dytox}.
%




\begin{figure*}[b]
   \centering
\includegraphics[width=0.97\textwidth]{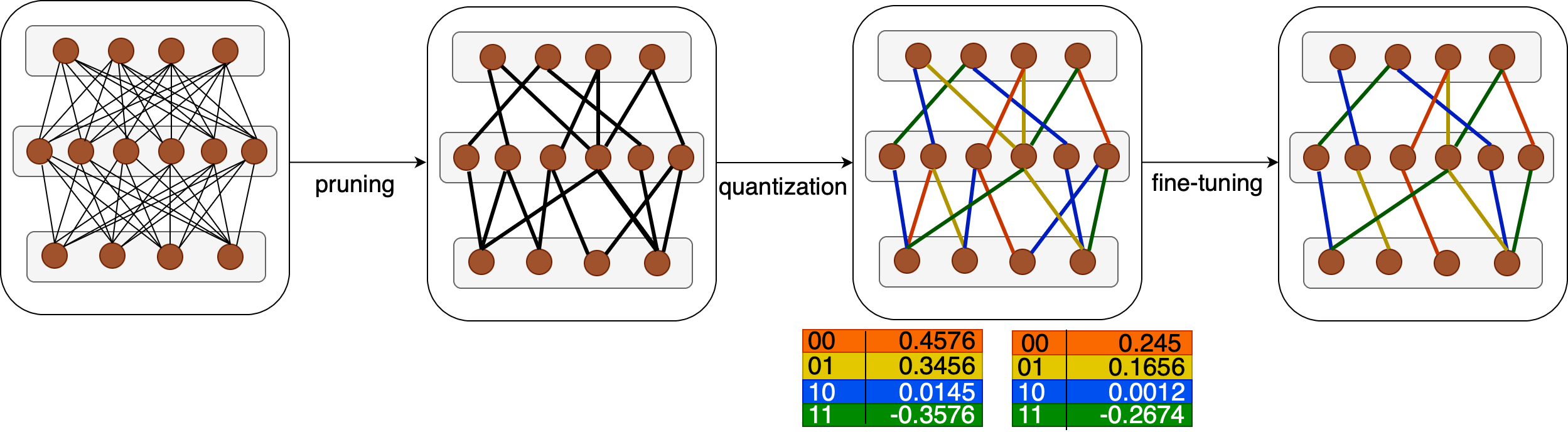}
\caption{Model architecture evolution with TinySubNetworks (TSN). To incorporate a new task, the model-agnostic continual learning workflow involves pruning, non-linear quantization (different colors represent different clusters from the codebook), and fine-tuning (which involves additional weight pruning without re-training).  
  }
    \label{fig:algorithm-architecture}
\end{figure*}

In this paper, we propose TinySubNets (TSN), a novel forget-free method for continual image classification that addresses these limitations by incorporating adaptive pruning with sparsity-level identification and adaptive non-linear weight quantization (Figure \ref{fig:algorithm-architecture}). 
To effectively exploit model sparsity, TSN incorporates pruning with different sparsity levels for each layer. 
Moreover, to deal with the issue of quickly saturating model capacity, TSN performs an adaptive quantization stage, where each task quantizes its weights via a stored codebook obtained via clustering. Quantized weights can be shared between tasks if their KL-divergence is low, or they can be separated into components, i.e. subsets of the available 32 bits, if their KL-divergence is high, resulting in totally independent memory banks with reduced bit-widths. 
Weight sharing is carried out through a trainable mask that automatically selects relevant weights for the current task from previous tasks during gradient descent learning, minimizing the bias associated with learning previous tasks. 
Thanks to these capabilities, our strategy allows model capacity to be drastically reduced while preserving model accuracy.



In summary, the contributions of our paper are as follows:
\begin{itemize}
\item{We propose a forget-free model-agnostic continual strategy that adapts model architectures via adaptive pruning, quantization, and fine-tuning stages.} 


\item{We devise an approach to seamlessly perform and combine quantization with weight sharing and replay memory for storing task samples, leading to an effective trade-off between model performance and capacity exploitation.}
\item{We showcase the positive impact of our proposed strategy on model performance through an extensive experimental analysis. Our results show that TSN can outperform state-of-the-art continual learning strategies with commonly adopted scenarios and datasets.}
\end{itemize}

In the following sections, we summarize related works in CL strategies and pruning methods. 
Then, we describe our proposed TSN method. We follow with a description of our experimental setup, and we present the results extracted in our experiments. Finally, we provide a summary of the results obtained and outline relevant directions for future work.

\section{Related Works}
\label{sec:background}
Continual learning methods are commonly categorized as replay-based, regularization-based, and architectural-based \cite{parisi2019continual}.
Replay-based methods usually store some of the experiences form the past episodes and replay them while training with data from new tasks \cite{parisi2019continual}. The most popular methods are GEM \cite{lopezpaz2017}, A-GEM \cite{chaudhry2019}, and GDumb \cite{prabhu2020}. 

On the other hand, regularization methods usually put constraints on the loss function to prevent purging knowledge of already learned patterns. Such approach is part of methods such as Synaptic Intelligence (SI) \cite{zenke2017}, LwF \cite{li2017learning}, and EWC \cite{kirk2017}.
The authors in \cite{liang2024loss} have recently proposed loss decoupling (LODE) for task-agnostic continual learning. LODE separates the two objectives for new tasks by decoupling the loss, allowing the model to assign different weights for different objectives, and achieving a better trade-off between stability and plasticity.
The work in \cite{mcdonnell2024ranpac} proposes a continual learning approach for pre-trained models. Since forgetting occurs during parameter updating, the authors exploit training-free random projectors and class-prototype accumulation. They inject a frozen random projection layer with nonlinear activation between the pre-trained model's feature representations and output head to capture interactions between features with expanded dimensionality, providing enhanced linear separability for class-prototype-based continual learning. 
Although regularization-based methods represent a viable way to realize continual learning, one major drawback is that introducing new tasks exacerbates the forgetting of previously learned tasks over time.


Architectural-based methods concentrate on the topology of the neural model trying to alter it or leverage the available capacity to prevent the model from forgetting. One of the most known methods is CWRStar \cite{lomonaco2019nicv2} that freezes all weights except the last layer that is trained to accommodate new tasks.
One of the key components of many architectural-based methods is pruning that allows to remove weights less important to already learned tasks and reuse them to accommodate new knowledge. For example, PackNet \cite{mallya2017} leverages pruning to divide the neural network into independent sub-networks assigned for specific tasks. This type of approach is known as forget-free, as having independent sub-networks ensure full protection from forgetting.
The DyTox algorithm \cite{douillard2022dytox} is a quite novel approach based on the Transformer architecture, which shows efficient usage of transfer learning in continual learning process. However, its main disadvantage is that the same architecture with the same model capacity is used for all CL scenarios.
Ada-QPacknet \cite{pietron2023ada}  adopts compression techniques like pruning and quantization as a viable approach to deal with model capacity. However, despite its efficiency on a number of common CL scenarios, this method does not support weight sharing between tasks. 
Moreover, its pruning step is based on lottery ticket search without leveraging gradient mask optimization and post-training fine-tuning. Another potential drawback is the lack of a mechanism to measure and exploit differences in task distributions.

\begin{figure*}[b]
   \centering
\includegraphics[width=0.995\textwidth,trim=28 5 6 8,clip]{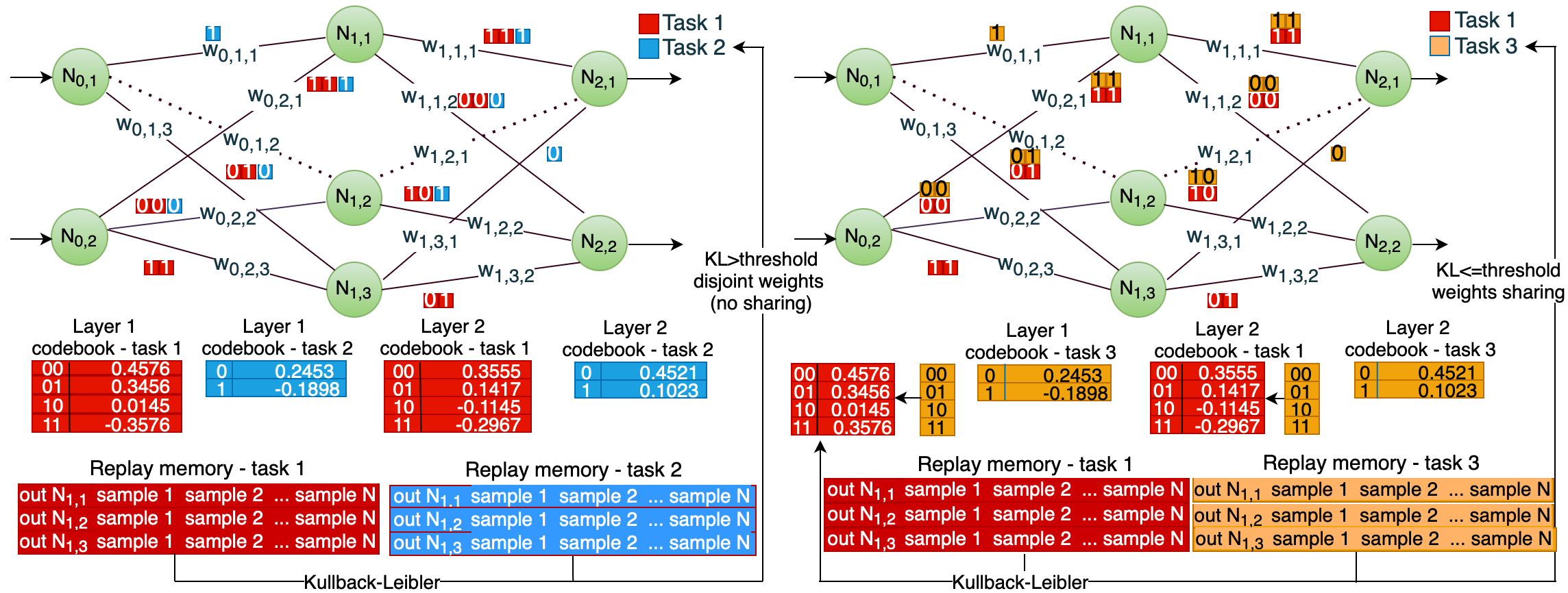}\\
       (a) \ \ \ \ \ \ \ \ \ \ \ \ \ \ \ \ \ \ \ \ \ \ \ \ \ \ \ \ \ \ \ \ \ \ \ \ \ \ \ \ \ \ \ \ \ \ \ \ \ \ \ \ \ \ \ \ \ \ \ \ \ \ \ \ \ \ \ \ \ \ \ \ \ \ \ \ \ \ \ \ \ \ \ \ \ \ \ \ \ \ \ \ \ (b)
\caption{TinySubNetworks (TSN) - scenario with physical connection sharing with disjoint values using a reduced bit-width format where weight values are not shared between tasks (a), and with value-based weight sharing where overlapping weight values are shared between tasks (b). Different colors represent different tasks.}
    \label{fig:algorithm}
\end{figure*}

A number of recently proposed CL works can be regarded as hybrid, which often provide a mixture of regularization and architectural approaches, with memory-based features.  
The work in \cite{jha2024npcl} proposes an NP-based approach with task-specific modules arranged in a hierarchical latent variable model, in which regularizers on the learned latent distributions are adopted to alleviate forgetting. Uncertainty estimation is supported to handle the task head/module inference challenge. 
%
A different approach is taken in \cite{madaan2023heterogeneous} where, inspired by the knowledge distillation framework, the authors devise a setting where a weaker model takes the role of a teacher, while a new stronger architecture acts as a student. To deal with limited data availability, they propose Quick Deep Inversion (QDI) \cite{madaan2023heterogeneous} to enhance distillation by recovering prior task visual features that support knowledge transfer. 
%
However, this work does not discuss model capacity, making it difficult to gauge the trade-off between accuracy and memory requirements.
A relevant method with memory-based features is Pro-KT \cite{li2024learning}, a prompt-enhanced knowledge transfer model to open-world continual learning, i.e. learning on the fly
with the goal of recognizing unknowns. The method is based on a prompt bank to encode and transfer task-generic and task-specific knowledge, and a task-aware open-set boundary to identify unknowns in the new tasks. 
Another example is Scaled Gradient Projection (SGP) \cite{saha2023continual} , which combines
 orthogonal gradient projections with scaled gradient
steps along the important gradient spaces for the past tasks.
The authors scale gradients in these spaces based on
their importance, and formulate an efficient way to compute it via singular value decomposition of input task representations.

Architectural and hybrid methods can be regarded among the most intriguing and sophisticated continual learning strategies. However, the major limitation of these methods appears when the capacity required to accommodate new tasks grows over time until exhaustion. Therefore, efficient capacity exploitation is of paramount importance and represents an open challenge in continual learning. 

\section{Method}
\label{sec:method}
Our TinySubNets continual learning strategy is a forget-free approach. The algorithm incorporates pruning and quantization with weight sharing, which allows us to drastically reduce model capacity.
TinySubNets can be run in two configurations: with and without weight sharing. 
Weights sharing between tasks is implemented by introducing a mask that automatically selects weights from previous tasks for the current task during gradient descent learning. This capability has the advantage to minimize the bias associated with learning previous tasks.
%
In the case of tasks with a significantly different distribution, the algorithm introduces the possibility of dividing quantized weights into memory banks.

The pruned model can be represented as $F_{\Theta}^{p} = (F_{\Theta}, M)$, where $F_{\Theta}$ is defined as:
\begin{equation}
F_{\Theta}(X) = {f_{\theta_L}(f_{\theta_{L-1}}...(f_{\theta_{0}}(X)))}.
\label{eq:model}
\end{equation}
$F_{\Theta}$ is the initial full model with a set of convolutional and fully-connected layers $f_{\theta_{i}}$, which are defined in the specified order. The $\Theta$ tensor is defined as: 
\begin{equation}
\Theta = \{\theta_{0}, \theta_{1},...,\theta_{L}\},
\label{eq:theta_}
\end{equation}
where $\Theta$ represents weights of all layers of the model, while $\theta_i$ contains weights from connections between layer $i$ and $i+1$. $w_{i, k, l} $ denotes a single weight between $k$-th neuron from layer $i$ and $l$-th neuron from layer $i+1$ (Figure \ref{fig:algorithm}).
\footnote{To simplify our notation, we assume that a single weight value is assigned to all tasks. In our method, multiple values for a single weight can be assigned to solve multiple tasks.}

A key aspect of the method is the adoption of masks, which denote assignments of weights to specific tasks. All masks are stored in a tensor $M$:
\begin{equation}
M = \{M^{0}, M^{1},...,M^{L}\}.
\label{eq:mask_}
\end{equation}
A single mask $M^i$ corresponds to weights between layer $i$ and $i+1$ and has the same size as $\theta_i$. Each entry $M^i_{k, l}$ in mask $M^i$ describes a set of tasks in which the weight between $k$-th neuron from layer $i$ and $l$-th neuron from layer $i+1$ is leveraged. This single entry is defined as:
\begin{equation}
M^i_{k, l} = \{t_1, t_2, \dots, t_n \},
\label{eq:mask__}
\end{equation}

where $t_p$ is a binary value indicating if weight is pruned or not by given task. The quantization generates bank, prefix, and key in the codebook to which a given weight is assigned for solving task $p$. 
Figure \ref{fig:algorithm} showcases a graphical representation of this process.
While the bank defines a subset of the 32-bit bandwidth of the weights, the prefix defines the specific codebook to which a weight is assigned in one bank.

\begin{algorithm}
\begin{algorithmic}[1]
\REQUIRE{$D = \{D_1, D_2, \dots, D_{T}$\} -- training data for tasks $1,2, \dots, T$}
\REQUIRE{$c$ -- initial capacity per task}
\REQUIRE{$\Theta$ -- model weights}
\REQUIRE{$M_{0}=0^{\Theta}$ -- initial binary mask}
\REQUIRE{$k$ -- frequency (num. of batches) to trigger adaptive quantization}
\REQUIRE{$\omega$ -- Kullback-Leibler threshold}
\REQUIRE{$bs$ -- batch size}
\REQUIRE{$s_{r}$ -- task replay memory size}
\STATE{Randomly initialized $\Theta$ and $s$} 
\FOR{$t = 0$ \TO $T$}
\STATE{$R_{t}$ $\gets$ $s_{r}$ from $D_t$ to replay memory}
\STATE{$p$ $\gets$ -1}
\FOR{$p=t-1$ \TO 0}
\STATE{$kl$ $\gets$ compute $D_{KL(R_{t}, R_{p})}$}
\IF{$kl$ < $\omega$}
\STATE{set weight sharing of task $t$ with task $p$}
\STATE{break}
\ENDIF
\ENDFOR
\FOR{$b = 0$ \TO $\frac{|D_{t}|}{bs}$}
\IF{$p$ $>$ -1}
\STATE{Obtain mask $M_{t}$ of the top-c\% weights at each layer sharing with task $p$ based on score $s$}
\ELSE
\STATE{Obtain new random mask $M_{t}$ of the the separate weight memory bank}
\ENDIF
\IF{$b \ \% \ k == 0$}
\STATE{$\phi_{t}$, $\Theta$, $K_{t}$ $\gets$ adaptive quantization $\Theta$ $\odot$ $M_{t}$} \{ \ \ See Appendix for Algorithm\ \ \}
\ELSE
\STATE{$\Theta$ $\gets$ $\Theta$ $\odot$ $M_{t}$}
\ENDIF
\STATE{compute $\mathcal{L}_{(x,y)}$}
\STATE{$\Theta$ $\gets$ $\Theta$-$\eta$ $\cdot$ ($\frac{\delta \mathcal{L}_{x,y}}{\delta \Theta}$ $\odot$ (1-$M_{t-1}$))}
\STATE{$s$ $\gets$ $s$-$\eta$ $\cdot$ ($\frac{\delta \mathcal{L}_{x,y}}{\delta s}$)}
\ENDFOR
\STATE{$M_{t}$ $\gets$ $M_{t-1}$ $\vee$ $M_{t}$ }
\STATE{$\Theta$, $M_{t}$ $\gets$ fine tuned pruning } \COMMENT{Algorithm \ref{alg:pruning}}
\STATE{$M$ $\gets$ $M$ $\cup$ $M_{t}$ }
\STATE{$K$ $\gets$ $K$ $\cup$ $K_{t}$ }
\ENDFOR
\RETURN{$\Theta$ -  weights, $M_{t}$ - mask, $K$ - tasks codebook}
\end{algorithmic}
\caption{TinySubNetworks (TSN) main algorithm}
\label{alg:main_algorithm}
\end{algorithm}

\begin{figure}[h!]
   \centering
  \includegraphics[width=0.45\textwidth]{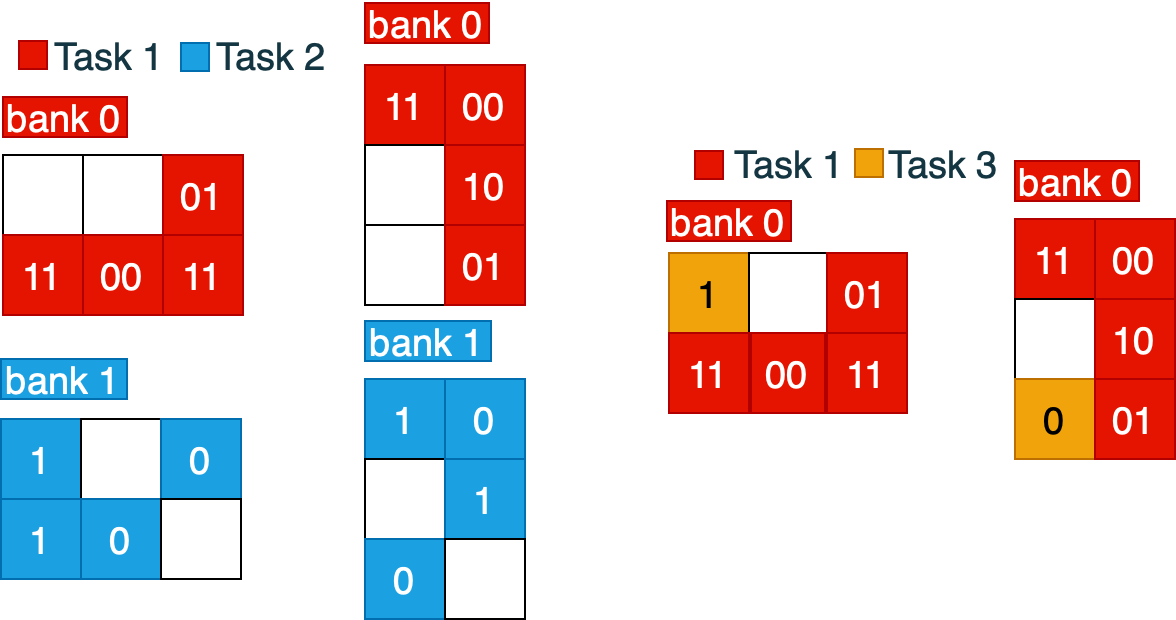} \\
    (a) \ \ \ \ \ \ \ \ \ \ \ \ \ \ \ \ \ \ \ \  \ \ \ \ \ \ \ \ \ \ \ \ \ \ \ \ \ \ \ \ \ \ \  \ \ \ \ \ \ \ \ \ \ (b)
\caption{TinySubNetworks - mask with connection sharing with disjoint values (a) and value-based weight sharing (b). Each subfigure represents the weight matrix of the model: rows (input size of the layer) by columns (output size of the layer). White boxes represent pruned weights. Different colors represent weight allocations for multiple tasks assigned to each mask. 
  }
    \label{fig:mask}
\end{figure}

A visual representation of learned masks is shown in Figure \ref{fig:mask}. Two types of weight sharing are supported: connection sharing with disjoint values (a), where separate masks with different values are learned for each task, and value-based weight sharing (b), where the same weight value is used for different tasks.

The sparsity level $\Upsilon_i$ describes how many weights out of all weights available in layer $i$ are not yet assigned to any task. It is defined as: 
\begin{equation}
\Upsilon_i =\frac{|M^{i}|-\sum_{k,l}^{{|M^{i}|}} (|M^{i}_{k, l}| > 0)}{|M^{i}|}
\label{eq:mask___}
\end{equation}

It is also possible to define the overall weighted sparsity of the model: 
\begin{equation}
    \Upsilon = \sum_i^L |\theta_i| \cdot \Upsilon_{i}
    \label{eq:w_sparsity}
\end{equation}






Additionally, masks for all tasks are encoded by Huffman algorithm to compress theirs size. The same methodology is used in \cite{kang2022}, see Appendix\footnote{\url{https://arxiv.org/abs/2412.10869}}.

The TSN main workflow is described in Algorithm \ref{alg:main_algorithm}.  The process starts with random initialization of values and score weights. In the following iterations, the model is trained with subsequent tasks. In the initial processing phase, the task is filled with replay memory i.e. a fixed number of input data samples for each class of a given task. Then, the divergence between the current task and all previous ones is calculated. 
We adopt the KL-divergence to measure the divergence between tasks, defined as:
%
%

%
\begin{equation}
D_{KL_{t_{i}, t_{j}}} = D_{KL(D_{t_{i}}, D_{t_{j}})}.
\end{equation}
Specifically, a low KL-divergence implies that quantized weights may be shared between tasks.
If the divergence is lower than the chosen threshold, the task with the minimum divergence is selected as the candidate for weight sharing. 
%
On the contrary, a high KL-divergence implies that tasks are significantly different. 
If the divergence is greater than the given threshold for all previous tasks, the current task allocates its own weight subspace by reserving a memory bank with a reduced number of bits  (Lines 12-16).

\begin{algorithm}[h]
\begin{algorithmic}[1]
\REQUIRE{$I$ -- fine tuning iterations}
\REQUIRE{$\Theta$ -- weights of the model}
\REQUIRE{$M_{t}$ -- current mask of the task}
\REQUIRE{$\Delta$ -- quant of sparsity change}
\REQUIRE{$A$, $\Upsilon$ -- initial accuracy and sparsity of the task}
\REQUIRE{$\alpha$, $\beta$ -- scaling factors}
\STATE{$\Theta$ $\gets$ initialize the task weights by $\Theta$ $\odot$ $M_{t}$}
\STATE{$M_{opt}$ $\gets$ $M_{t}$}
\STATE{$\gamma_{opt}$ $\gets$ $\alpha \cdot A+\beta \cdot \Upsilon$}
\FOR{$i = 0$ \TO $I$}


%


\STATE{$l$ $\gets$ choose layer from 0 to $L$}
\STATE{$M'$ $\gets$ increment mask of layer $l$ - $M_{l}$} by $\Delta$
\STATE{$\Theta'$ = $\Theta$ $\odot$ $M'$} 
\STATE{$A' \gets$ accuracy of $\Theta \odot$ 
 $M'$ on task $t$}
\STATE{$\Upsilon' \gets$ compute sparsity for $M'$ (eq: \ref{eq:w_sparsity})}
\STATE{$\gamma$ $\gets$ $\alpha \cdot A'+\beta \cdot \Upsilon'$}
\IF{$\gamma$ > $\gamma_{opt}$}
\STATE{$M_{opt}$ $\gets$ $M'$}
\STATE{$\gamma_{opt}$ $\gets$ $\gamma$}
\ELSE
\STATE{$M_{l}$ $\gets$ decrement mask $M_{l}$ of layer $l$ by $\Delta$}
\ENDIF



\ENDFOR 

\STATE{$\Theta' \gets \Theta \odot M_{opt}$}
\RETURN{$M_{opt}, \Theta' $}
\end{algorithmic}
\caption{Fine tuned pruning for task $t$}
\label{alg:pruning}
\end{algorithm}

Then, the training process begins. In the feed-forward process, weights are multiplied by the current mask which is set based on weight scores $s$. 
If a batch is a multiple of the $k$ parameter, then the adaptive quantization algorithm is also executed (for additional details see Appendix). Then, the loss is computed, and the weights are updated:
{
\small
\begin{equation}
\mathcal{L}_{(x,y)}=\frac{1}{\left|B\right|} \sum_{b=1}^{\left|B\right|} \Bigg(\mathcal{L}_{ce}(f_{\theta}(x_{b}),y_{b})+\alpha \sum_{i=0}^{L} \mathcal{L}_{mse}(\overline{f_{\theta_{i}}},f_{\theta_{i}})\Bigg),
\end{equation}
}
where $\mathcal{L}_{ce}$ is the conventional cross-entropy loss iteratively computed over batches $B$, and $\mathcal{L}_{mse}$ measures the differences in layer representations computed between the full model $f_{\theta_{i}}$ and the compressed model $\overline{f_{\theta_{i}}}$ which features pruning and quantization stages.
After the learning process for a given task is completed, the global consolidated mask is updated. Next, pruning optimization is performed on the validation set. In the last step, the global mask and codebook are updated. 

In Algorithm \ref{alg:pruning} the post-pruning without retraining process is presented. This process has a very limited computational cost (see execution times in Appendix), and its goal is to check the possibility to slightly increase the sparsity level if it does not impact accuracy, without triggering model retraining. 
The algorithm starts by setting the following values: weights multiplied by the input mask (mask achieved after training stage), the optimal mask (which is initially equal to the input mask), and the  $\gamma_{opt}$ variable, which constitutes the fitness function of the optimization process. Its value is obtained as the  the sum of two components:  the accuracy value multiplied by the $\alpha$ coefficient, and the sparsity value multiplied by the $\beta$ coefficient (in all experiments $\alpha$ is set to 0.95 and $\beta$ to 0.05). 

Subsequently, the main loop of the algorithm is executed. In the next search iterations, the model layer is selected as a candidate for increasing sparsity. Then the mask is incremented (the number of zeros is increased). In the next stage, the weights are modified according to the changed mask. Accuracy is calculated for the changed weights and a new sparsity value is set. Based on them, the gamma value is calculated. If $\gamma$ is greater than $\gamma_{opt}$, the modified mask becomes the current mask of the given layer and the $\gamma$ value becomes $\gamma_{opt}$. Otherwise, the layer mask remains unchanged. 

Moreover, in addition to the main TSN method described above, we also introduce TSN-wr (TSN without replay). In this case, weight sharing is based on pruning instead of replay buffer. While only not yet used weights are altered during learning a new task, the other weights are still used in the forward pass. When pruning weights after learning a new task, some of the weights already assigned to another task may still be assigned to a new task as well.
The TSN-wr follows the same algorithm as TSN (Algorithm \ref{alg:main_algorithm}) except for lines 3-10, which are not included in TSN-wr execution. While, intuitively, this may lead to decreased performance in terms of accuracy, it should also allow to reduce the used memory due to lack of replay-buffer.

\section{Results and Discussion}
\label{sec:results}

\textbf{Datasets:} Our main experiments were performed with three popular and commonly adopted CL scenarios: Permuted MNIST (p-MNIST) \cite{lecun1998mnist}, split CIFAR100 (s-CIFAR100) \cite{Krizhevsky2009LearningML}, and 5 datasets \cite{ebrahimi2020adversarial}, a task-incremental scenario consisting of MNIST, SVHN, FashionMNIST, CIFAR10, not-MNIST, TinyImagenet \cite{le2015tiny}, and Imagenet100 \cite{russakovsky2015imagenet}. The p-MNIST scenario consists of 10 tasks with randomly permuted pixels of the original MNIST (10 tasks with 10 classes each). s-CIFAR100 is divided into 10 tasks with 10 classes each. The 5 datasets scenario is a sequence of 5 tasks with different datasets,  each with 10 classes.   
The TinyImagenet scenario consists of 40 tasks with 5 randomly sampled classes each.
Finally, the Imagenet100 scenario consists of 10 tasks with 10 randomly sampled classes each.

\textbf{Experimental Setup:} 
%
%
%
The adopted models are a two-layer neural network with fully connected layers (p-MNIST), reduced AlexNet (s-CIFAR100) \cite{saha2020gradient}, Resnet-18 (5 datasets and Imagenet100), TinyNet (TinyImagenet) in accordance with model backbones used in the WSN paper \cite{kang2022}. 
For weight initialization, we adopt the Xavier initializer. Experiments are executed on a workstation equipped with an NVIDIA A100 GPU. Our experiments involve 5 complete runs for each strategy. The training and testing execution times are reported in the Appendix\footnote{\url{https://arxiv.org/abs/2412.10869}}. 
The code of the method is available at the following public repository: \url{https://github.com/lifelonglab/tinysubnets}. 


\textbf{Metrics:} We adopt the most standard definition of lifelong Accuracy following the description in \cite{diaz2018don}.
Moreover, we also compute the capacity (total memory occupied by each task) inspired by the capacity computation in \cite{kang2022}. The capacity can be regarded as the sum of three components, and defined as follows: 
\begin{align*}
\scriptsize
    CAP_t = \sum_{i}^{L}(1-{\Upsilon_{i}})\cdot|\theta_i|\cdot b + |L| \cdot 2^b \cdot (32 + b) + \\
    \sum_{i}^{L}(1-\Upsilon_{i}) \cdot |M^i|  
\end{align*}

The first one describes the number of the weights after the pruning multiplied by the sum of bit-width, prefix and bank identifier of the codebook $b$. The second one describes the codebook size. The third one includes all masks' capacity. 

\textbf{Hyperparameters:} The hyperparameters used in our experiments were set up as follows: 
\begin{itemize}
\item \textbf{Adaptive learning rate} -- for p-MNIST it starts from $3e-1$ and ends in $1e-4$, for CIFAR100 and 5 datasets it starts from $1e-2$ and ends with $1e-4$, for the rest of the scenarios is between $1e-3$ and $1e-5$ 
\item \textbf{Number of epochs per task} -- 200 epochs per task for each scenario
\item \textbf{Batch size} -- CIFAR100 - 2048, Imagenet100 - 32, 5 datasets, p-MNIST and TinyImagenet - 256
\item \textbf{Fine tuning parameters} - 50 iterations for each scenario, $\alpha$ - 0.95, $\beta$ - 0.95, 
\item \textbf{Kullback-Leibler threshold} - set empirically to have max two memory banks
\item \textbf{Initial capacity per task} - 0.55 for CIFAR100, 0.5 for the rest of the scenarios
\item \textbf{Frequency} (num. of batches) to trigger adaptive quantization - three times per epoch for each scenario
\item \textbf{Task replay memory size} - 50 samples per task (only 15 samples per task for TinyImagenet),
\item \textbf{Quantity of sparsity change} - 0.01 for each continual learning benchmark
\end{itemize}
The SGD optimizer was used for p-MNIST dataset. For the other scenarios the ADAM optimizer was adopted.

\subsection{Comparative Studies}
Table \ref{table:CL_accuracy} presents the results of our experiments in terms of accuracy for two TSN variants: TSN without replay memory (TSN-wr) and TSN mixed (TSN with replay and fine tuning). We also present the results of the most popular standard CL methods and upper-bounds (Naive, CWRStar, SI, Replay, Cumulative), as well as three  forget-free architectural methods: Packnet, WSN, and Ada-QPacknet.

Moreover, we also present capacity comparison for pruning-based methods in Table \ref{table:CL_capacity} where results for TSN without replay and fine tuning (TSN-wr-wpp) version are added.
Additional results such as backward, forward transfer, sparsity levels, bit-widths and masks compression ratios achieved by Huffman algorithm are reported in the external Appendix.


\vspace{5pt}

\textbf{Accuracy-Capacity Tradeoff:}
As we can observe, TSN-wr allows us to significantly reduce capacity in comparison to other pruning methods ($22.65\%$ vs $77.73\%$ for p-MNIST, $17.62\%$ vs $78.6\%$ for s-CIFAR100, $24.68\%$ vs $33.7\%$ for 5 datasets, and $32.15\%$ vs $48.65\%$ for TinyImagenet). 

At the same time, TSN-wr presents an accuracy that is just slightly worse than the best-performing method ($96.63\%$ vs $97.14\%$ for p-MNIST, $75.21\%$ vs $77.27\%$ for s-CIFAR100, $91.8\%$ vs $94.1\%$ for 5 datasets, and $79.81\%$ vs $80.10\%$ for TinyImagenet). 
Results obtained for the TSN method were obtained setting the Kullbach-Leibler divergence threshold empirically to have two memory banks at most, in order to balance memory consumption and accuracy.
It is worth mentioning that Algorithm \ref{alg:pruning} resulted, on average, in an improvement from $1\%$ (p-MNIST) to $6\%$ (5 datasets) in capacity utilization (TSN-wr-wpp), with a negligible drop in accuracy (less than $1\%$, see Appendix). 
%
Overall, it is possible to observe that TSN-wr provides the best trade-off between accuracy and capacity. 
%
On the other hand, TSN achieves SOTA results on p-MNIST ($97.14\%$), and TinyImageNet ($80.10\%$), and s-CIFAR100 ($77.27\%$), at the cost of a higher capacity when compared to TSN-wr ($93.6\%$ vs $32.15$ on TinyImagenet and $37.5\%$ vs $22.65\%$ on p-MNIST and $41.92\%$ vs $17.62\%$ on s-CIFAR100). 

%

\vspace{5pt}

\textbf{Impact of Weight Sharing:} To assess the impact of weight sharing, it is relevant to compare our approach which involves weight sharing, with architectural strategies without weight sharing, such as Ada-QPacknet. Our results show that weight sharing is beneficial in scenarios with low task heterogeneity, where task similarity can be profitably exploited (e.g. p-MNIST; s-CIFAR100). 
%
%
Quite interestingly, the situation is different in scenarios where tasks are more heterogeneous, such as 5 datasets. In this scenario, strategies without weight sharing achieve the best accuracy ($94.1\%$), a $2.3\%$ improvement over the replay variant of TSN.
With TinyImagenet, we observe that TSN-wr outperforms Ada-QPacknet ($79.81\%$ vs. $71.9\%$). Despite the large number of classes in this dataset, we argue that TinyImagenet has a limited class representation, i.e. number of images per class, which makes the scenario less representative in terms of complexity than 5 datasets.

\begin{table}
\setlength{\tabcolsep}{1pt}
\small
\centering
\begin{tabular}{|c|c|c|c|c|} 
\hline
  & \textbf{p-MNIST} & \textbf{s-CIFAR100} & \textbf{5 datasets} & \textbf{Tiny} \\ 
\textbf{}   & \textbf{} & \textbf{} & \textbf{} & \textbf{Imagenet} \\ 
\hline
\textbf{Packnet}    &   96.38\%                      &            81.0\%       &         82.86\%   & 188.67\%                  \\ 
\textbf{WSN}        &         77.73\%                  &       99.13\%                &           86.10\%  &   48.65\%                    \\ 
\textbf{Ada-QPacknet}      &     81.25\%                 &   78.6\%              &   33.7\%            &     112.5\%     \\ 
\textbf{TSN-wr}      &      \textbf{22.65\%} 
&    \textbf{17.62\%} 
&     \textbf{24.68\%} 
&     \textbf{32.15\%} \\
\textbf{TSN-wr-wpp}      &   23.41\%   &   18.75\%    &    30.62\% 
&  36.33\% \\
\textbf{TSN}      &     37.5\%*
&      41.92\%     
&     40.08\% 
&  93.6\%  
\\

\hline
\end{tabular}
\caption{Capacity comparison for pruning-based methods (* - in case of p-MNIST two memory banks without replay memory).}
\label{table:CL_capacity}
\end{table}

\vspace{5pt}

\textbf{{Continual Learning} Strategies on ImageNet100}: Results in Table \ref{table:CL_capacity_new_competitors} present several SOTA architectural and replay methods on the ImageNet100 scenario.  
%
DER \cite{buzzega2020} has the highest parameter count at $112.27$ million, significantly larger than the other methods. Despite its high complexity, DER achieved an accuracy of $75.36\%$, which is slightly lower than DyTox.
Ada-QPacknet has 11.5 million parameters, and achieved an accuracy of $72.26\%$, which is the lowest among the considered methods. 
Overall, we can observe that TSN-wr achieves the best results in terms of both parameters number and accuracy. 
With only $2.47$ million parameters, TSN achieved a notable accuracy of $77.16\%$, which is particularly impressive, and suggests that TSN is highly efficient and effective.
This result suggests that utilizing more capacity (as in DER or ADA-QPacknet) does not necessarily translate to a higher accuracy in complex scenarios.

\vspace{5pt}

\textbf{Ablation Studies:} We perform two ablation studies to investigate the memory capacity requirements for weights, codebooks, and masks across scenarios in weight-sharing mode. 
Our study suggests that while weights and masks are the primary consumers of memory, the use of codebooks introduces minimal additional memory requirements. 
The second ablation study evaluates the accuracy of models using different bit widths. 

Our study shows that while reducing the bit width can help in compression, it often comes at the cost of accuracy. The impact varies by dataset, with larger and more complex datasets like TinyImagenet experiencing a more pronounced drop in accuracy at lower bit widths. For CIFAR100, a 4-bit width seems optimal, balancing efficiency and performance. This suggests that the choice of bit width should be carefully considered based on the specific dataset and accuracy requirements. Quantitative detailed experiments for the two ablation studies are reported in the Appendix. 
In all experiments variance in accuracy does not exceed 0.04. All experiments were run five times.

\begin{table}[htbp]
\setlength{\tabcolsep}{1pt}
\small
\centering
\begin{tabular}{|c|c|c|c|c|} 
\hline
\textbf{}   & \textbf{p-MNIST} & \textbf{s-CIFAR100} & \textbf{5 datasets} &  \textbf{TinyImagenet} \\ 
\hline

\textbf{Naive}      & 60.12                     & 17.32                  & 33.08  &                         20.27      \\ 
\textbf{CWRStar}    & 31.31                     & 20.84                  & 36,16     &                       24.00       \\ 
\textbf{SI}         & 57.32                     & 19.54                  & 29.42      &                   20.51        \\ 
\textbf{Replay}     & 62.22                     & 19.60                  & 55.24       &                23.14         \\ 
\textbf{Cumulative} & 96.45                     & 36.52                  & 84.44        & 27.53                        \\ 
\textbf{Packnet}    &  96.31                         & 72,57                  &      92.59   & 55.46 \\ 
\textbf{WSN}        &      96.41                     &       76.38              &    93.41 &  71.96  \\ 
\textbf{AdaQPacknet} &         97.1            &        74.1             &       \textbf{94.1}   & 71.9 \\
\textbf{TSN-wr} &           96.63             &    75.21 
&    91.80  &   79.81 \\
\textbf{TSN} &       \textbf{97.14}               &            \textbf{77.27}     &   93.76   &  \textbf{80.10}  \\
\hline
\end{tabular}
\caption{Comparative results in terms of average accuracy for all CL strategies with different CL scenarios (4-bit quantization): Permuted MNIST (p-MNIST), split-CIFAR100 (s-CIFAR100), and 5 datasets (5 trials, std < 0.002 ).}
\label{table:CL_accuracy}
\end{table}

\begin{table}[h!]
\small
\centering
\begin{tabular}{|c|c|c|} 
\hline
\textbf{Methods}   & \textbf{Parameters} & \textbf{Accuracy}  \\ 
\hline
\textbf{DyTox}        &   10.73M                       &      75.54                                         \\ 
\textbf{DER}        &      112.27M                    &          75.36                                     \\ 
\textbf{Ada-QPacknet}      &     11.5M                 &   72.26                       \\ 
\textbf{TSN-wr}      &        \textbf{2.47M}             &     77.16                    \\ 
\textbf{TSN}      &    4.94M   &    \textbf{77.86}      \\ 
\hline
\end{tabular}
\caption{Comparison with SOTA architectural and replay methods on Imagenet100 (10/10) (5 trials, std < 0.002 ).}
\label{table:CL_capacity_new_competitors}
\end{table}

\vspace{10pt}

\textbf{Energy Efficiency:} An important aspect of our method is its limited energy usage. This aspect is a key factor for model deployment in low-resourced devices such as mobile phones and IoT, and it has been recognized as crucial for Green AI \cite{schwartz2020green}, which promotes environmentally sustainable models with low carbon footprint. To demonstrate this capability, we compute the theoretical complexity of our proposed method in terms of floating point operations per second (FLOPS) with different quantization levels. 

\begin{figure}[h!]
   \centering
\includegraphics[width=0.45\textwidth,trim=35 5 35 30,clip]{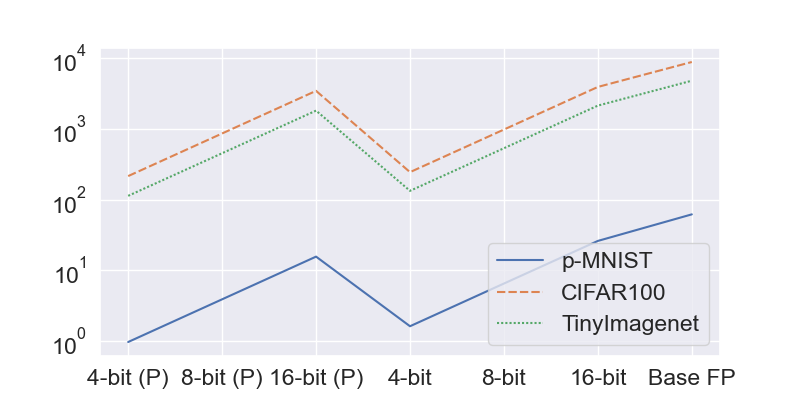}
  \caption{FLOPS with different quantization levels in different scenarios. P denotes the  method with pruning. Baseline FP denotes the full precision of weights (no compression).}
    \label{fig:flops}
\end{figure}

To this end, FLOPS calculations for GPU is carried out via arithmetic multiplication based on the size and number of convolutional and fully connected layers in model architectures. 
Results in Figure \ref{fig:flops} show that 16-bit and 8-bit quantized models provide a significant reduction in terms of FLOPS. This is a remarkable result considering the negligible drop in accuracy presented by compressed models. \footnote{The number of FLOPS and accuracy for each bit-width are presented in our external Appendix: \url{https://arxiv.org/abs/2412.10869}}.    



\section{Conclusions}
\label{sec:conclusions}
This paper introduces the TinySubNets (TSN) method, a forget-free continual learning strategy that provides an effective trade-off between model performance and capacity exploitation. TSN ensures this by adapting model architecture through pruning, adaptive quantization, and fine-tuning. Moreover, TSN is a model-agnostic approach and can be adapted to support any neural network architecture.
Our extensive experimental evaluation includes the most popular continual learning benchmarks and models, as well as multiple other forget-free and architectural methods. We observe that TSN is able to significantly reduce the used model capacity while keeping a performance similar (or even better) to its competitors. This reduction may be significant in extending the longevity of the models, allowing them to learn more tasks.
In future work, we will attempt to showcase the benefits of quantization performance while using specialized hardware accelerators. Moreover, we will also investigate the impact of decreasing the network size on the performance of architectural continual learning methods.


\newpage

\section{Acknowledgments}
We acknowledge the support of American University’s National Science Foundation (NSF) funded ADVANCE AU Career Development Mini-Grants, the Polish Ministry of Science and Higher Education, and AGH University (program "Excellence initiative - research university").

\bibliography{aaai25}  






\newpage

\section{Appendix/Supplementary Material}

\subsection{Adaptive quantization and non-linear quantization}

Algorithm \ref{alg:iter_quant} showcases the adaptive quantization process. The procedure begins by applying non-linear quantization to the task-pruned model using minimal values for the parameters $\omega$ (number of centroids) and $\psi$ (bit-width) (line 1). This initial step utilizes the non-linear quantization algorithm described in Algorithm \ref{alg:quant}.
Following the quantization, the model's accuracy is evaluated (line 2). If the accuracy drop exceeds a predefined threshold $\delta$ (line 3), an iterative process is initiated to incrementally increase the bit-width (lines 4-7). This adjustment continues until the accuracy loss is within acceptable limits.

\begin{algorithm}[h]
\begin{algorithmic}[1]
\REQUIRE{$\psi$ -- Initial bit-width}
\REQUIRE{$Q_q$ -- Model accuracy}
\REQUIRE{$\Theta$ -- Pruned model}
\REQUIRE{$\delta$ -- Maximum loss in accuracy}
\STATE{$\Theta_q, K$ = nonlinear\_quantization($\psi, \Theta$)} \COMMENT{see Algorithm \ref{alg:quant}}
\STATE{$Q_q$ = acc($F$, $\Theta_q)$}
\WHILE{$Q_q$ $<$ $Q$ - $\delta$}
\STATE{$\psi$ = $\psi$ + 1}
\STATE{$\Theta_q, K$ = nonlinear\_quantization($\psi, \Theta$)}
\STATE{$Q_q$ = acc($F$, $\Theta_q)$}
\ENDWHILE
\RETURN{$\psi$ -- optimal bit-width for the task, $\Theta_q$ -- quantized model, $K$ -- codebook}
\caption{Adaptive Quantization for task $t$.}
\label{alg:iter_quant}
\end{algorithmic}
\end{algorithm}

\begin{algorithm}[h]
\begin{algorithmic}[1]
\REQUIRE{$\psi$ -- desired bit-width}
\REQUIRE{$\Theta$ -- weights of a model}
\STATE{$\omega \gets 2^{\psi}$} \COMMENT{Number of centroids}
\STATE{$K \gets [K_0, K_1, \dots, K_{|\Theta|}]; K_i = \{ \}$} \COMMENT{Empty codebook}
\STATE{$\Theta' \gets$ copy of $\Theta$}
\FOR{$\theta_i$ $\textbf{in}$ $\Theta$}
\STATE{$C, A \gets$ KMeans($\theta_i$, $\omega$)} \COMMENT{$C$: centroids, $A$: weight-to-centroid-index assignment}
\FOR{$\theta_{i, j} \in \theta_i$}
\STATE{$\theta'_{i, j} \gets A[\theta_{i, j}]$} \COMMENT{Use centroid index instead of weight value}
\ENDFOR
\FOR{$c_k \in C$}
\STATE{$K_i[k] \gets c_k$} \COMMENT{Building layer-wise codebook}
\ENDFOR
\ENDFOR
\RETURN{$\Theta', K$}
\caption{Non-linear quantization}
\label{alg:quant}
\end{algorithmic}
\end{algorithm}

This adaptive approach ensures that the quantized model maintains a performance close to the original model while optimizing memory usage. By incrementally increasing the bit-width only when necessary, the algorithm finds a balance between model size and accuracy, making it effective for scenarios with strict memory constraints.

Algorithm \ref{alg:quant} presents our adopted non-linear quantization approach, which was inspired by \cite{pietronCANDAR}.
The process begins with identifying the number of centroids $\omega$ and creating an empty codebook $K$ to store the mapping between codes and weight values for each layer (lines 1-2). Additionally, we generate a copy of the model $\Theta$, where the weights will contain codes from the codebook (centroid indices) instead of actual values.

The next step involves clustering the weight values of each layer $\theta_i$ using the K-Means clustering algorithm, with a specified number of clusters ($\omega$) (line 5). This step produces two outputs: a list of centroids $C$ and a list of weight-to-centroid-index assignments $A$. Each weight is then assigned its corresponding centroid index (lines 6-8).

To maintain the mapping between centroid indices and centroids, a codebook is constructed for each layer (lines 9-11). The algorithm ultimately returns the model copy with quantized weights, now represented as codes from the codebook, and the codebook that is essential for converting the codes back to their actual weight values.

It is important to note that the number of centroids is directly related to the desired bit-width (as indicated in line 1). Fewer centroids result in reduced memory requirements for storing the weights, as each weight is stored as an index of its closest centroid. This approach ensures that memory usage is minimized while retaining the essential information needed for the model's performance.

\subsection{Formal details}
Pruning is defined as a function:
\begin{equation}
P : \theta_{i} \rightarrow {\theta_{i}'},
\theta_{i}' = M_{i} \odot \theta_{i}
\end{equation}
where: $M_{i}$ is a binary matrix with bits assigned to single weights 
\begin{equation}
\sum_{m \in M_{i}} M_{i} = \Upsilon_{i}
\end{equation}
\newline
Quantization assigns centroids to weights:
\begin{equation}
Q : \theta_{i} \rightarrow Y_{K_{i}}, Y_{K_{i}} \equiv \min \forall_{Y_{K_{i}}} |Y_{K_{i}} - \theta_{i}|
\end{equation}
\newline
\begin{equation}
K_{i} : X \rightarrow Y, X \in \mathbb{Z} \land Y \in \mathbb{R}
\end{equation}
\begin{equation}
|X| = 2^{\phi_{t}}
\end{equation}
The equation for accuracy loss can be defined as:
\begin{equation}
\mathcal{L}_{(x,y)} = \mathcal{L}_{ce}(Y,F_{Q(P(\Theta))}) + \sum_{i=0}^{L} \mathcal{L}_{mse}(\overline{f_{Q(P(\theta_{i}))}}, f_{Q(P(\theta_{i}))})
\end{equation}
\newline 
Kullback-Leibler for computing tasks divergence is defined as:
\begin{equation}
D_{KL}(D_{t}||D_{t-1}) = P(D_{t})\cdot \frac{P(D_{t})}{Q(D_{t-1})}
\end{equation}
\newline

\subsection{Huffman encoding and decoding}

Each tasks binary mask is represented as binary matrix. Each value indicates if weight is used (value 1) or not used by the task (value 0). It is just sparse matrix with nonzero 1 values. Therefore each mask can be compressed. In presented approach mask is encoded using huffman algorithm. The most frequent sub-sequences with their probabilities are are extracted. 
Then mask is represented as a tree and the dictionary in which mapping is stored (pairs with sequence in original mask and its corresponding code). When mask is loaded the decoding process is run to receive original mask. 

\subsection{Decoding quantized weights}

The quantized weights are encoded in following manner (from fig.2):
\begin{itemize}
    \item bank
    \item prefix (which indicates to which task weight belongs to)
    \item quantized weight value in the codebook
\end{itemize}

\textbf{Example:} 0 00 11 represents the value -0.3576 (weight $w_{0,1,3}$ in layer 1 and task 1, see fig.2a)

\subsection{Case study} 
In this subsection we describe more in detail an example of how TinySubNetworks manages memory capacity, as shown in Figure \ref{fig:algorithm} (with weight sharing, without replay memory, task 1, 2 and 3 share the weights, TSN-wr):
\begin{itemize}
\item Original model has capacity in bits:
{$CAP_{L_{1}}$ = $6 \cdot 32b$, $CAP_{L_{2}}$ = $6 \cdot 32b$, $\rightarrow$ $384b$}. 
\item After pruning of task 1 capacity: $4$ weights in first layer and $4$ in second layer, after pruning and quantization of task 1: $CAP_{t1}=4\cdot 2b + 4\cdot 2b + 136b \rightarrow 152b$. 
\item After pruning of task 2 capacity: it adds $1$ weight in first layer and $1$ weight in second layer, after pruning and quantization of task 2: $CAP_{t2} = 152b + 2*1b + 66b \rightarrow 220b $. 
\item Task 3 uses weights added by second task and shares some subset of weights from task 1: $CAP_{t3}=220b$.
\end{itemize}


\subsection{Hyperparameter optimization}

In regards to hyperparameter optimization and overfitting,
we resorted to basic heuristics for suitable values for most hyperparameters.
Regarding the adaptive pruning process, after the learning process for a given task is completed, the global consolidated mask is updated, and  codebook are updated. This process leverages a validation set in order to prevent overfitting.
Moreover, sparsity levels in Algorithm 2 are optimized resorting to a greedy non-gradient fine-tuning approach, which does not incur in overfitting issues.

\subsection{Additional results}
In this subsection, we present different results that highlight specific aspects of our method. 
Results in Table \ref{table:capacity_codebooks_masks_weights} show that the capacity occupation in different scenarios for the compressed model. Masks and weights are a small portion of the original model (after pruning and quantization). Size of the codebooks is negligible.

In Table \ref{table:sparsity} sparsity before and after fine tuning is shown. In last column the optimal bit-width is reported.
Results in Table \ref{tab:appendix-accuracy-bit-width} show that a significantly high model accuracy can be achieved even with a very small number of bits for the quantized model (between 3 and 5 bits).
%

Results in Table \ref{tab:time} report the execution time for each scenario (training, and fine-tuning), which depends on the specific input size of each dataset, the model backbone, and the batch size. An interesting result is that, while CIFAR100 is a larger dataset than p-MNIST, its execution time is less due to the specific batch size configuration used in our experiments ($2048$ vs. $256$).

A different view on capacity is shown in results in Table \ref{tab:appendix-capacity-post-pruning-replay} (please note that results with replay memory include the memory occupation for the additional memory bank for each task - see Figure \ref{fig:algorithm}). Overall, we observe that post-pruning can reduce the memory footprint of the model. Its impact is particularly high with 5 Datasets (from $30.62\%$ to $24.68\%$). We can also observe that replay memory requires a consistent amount of capacity.  
%

Results in Tables \ref{tab:appendix-flops} and \ref{tab:flops_acc} allow us to highlight that the reduction in FLOPs provided by compression incurs a negligible accuracy loss. For instance, in the TinyImagenet scenario, 8 bit quantization allows to reduce from 4 775M FLOPs to 530.5M (or 450.9M with the pruning variant) at just 0.27 in accuracy drop.   
In Table \ref{tab:flops_acc} the accuracy after centroids quantization and activations quantization is described (in <4b, 4b>, <8b, 8b> and <16b, 16b> formats). It can be observed that in <8b, 8b> and <16b, 16b> configurations the accuracy is still at the same level. In case of <4b, 4b> there is a drop in accuracy. All these configurations can reduce number of FLOPs as it is shown in fig. 4. 
\textbf{Note:} Floating-point operations <flp, flp> have approximately 9× to 10× of the complexity of <8b, 8b> since there is 3× of the number of bits to represent the mantissas plus the remaining work required to handle exponents.

Forward and Backward Transfer are relevant metrics in continual learning. Results in Table \ref{tab:appendix-forward-backward} and Figures \ref{fig:heatmap-cifar}-\ref{fig:heatmap-imagenet} show zero Backward Transfer for all scenarios, which is expected given the forget-free capabilities of the proposed method. As for Forward Transfer, values should be compared to the performance of a random classifier considering the number of classes in each dataset. Overall, we observe that Forward Transfer performance is close to or slightly above random. However, a proper evaluation of this metric is cumbersome in a task-incremental setting, given that our method requires knowledge of the task identifier to create masks and the specific sub-network for each task.  

Heatmaps in Figures \ref{fig:heatmap-cifar}-\ref{fig:heatmap-imagenet} provide additional disaggregated information, showing model performance on each task. It can be observed that the performance varies significantly depending on the task (e.g., on Imagenet-100, the performance varies from $68.9\%$ in task 5 to $83.8\%$ in task 7). The heatmaps further highlight the forget-free capabilities of the method, as shown by the performance on each task which is constant and preserved throughout the entire scenario. 

\subsection{Comparison with WSN and Ada-Q-Packnet}

While we recognize that our proposed approach is inspired by the best features of WSN \cite{kang2022} (weight sharing) and Ada-QPacknet \cite{pietron2023ada} (pruning and quantization), we did not simply adopt these features. Instead, we developed more effective versions of the same, and devised our own method to synergically combine them.  

While Ada-QPacknet \cite{pietron2023ada} adopts compression techniques like pruning and quantization to deal with model capacity, it presents some significant differences with respect to our proposed approach. 

First, Ada-QPacknet does not support weight sharing between tasks. This is a significant drawback in the presence of scenarios with high task similarity (similar data distribution), since it incurs in an inefficient use of memory. 

Second, our approach to pruning and quantization is different. Specifically, instead of a lottery ticket search as in Ada-QPacknet, our pruning step involves gradient mask optimization and post-training fine-tuning (without gradient), which is also unique when compared to Winning SubNetworks (WSN).

Third, our approach leverages the Kullbach-Leibler Divergence to measure the differences in task distributions, which is a missing feature in both Ada-QPacknet and WSN.

\subsection{Quantitative comparison with NPCL and QDI}

Regarding NPCL \cite{jha2024npcl}, the authors leverage a ResNet-18 model backbone across all class incremental experiments and two-layer fully connected network for task incremental (p-MNIST). In contrast, we adopt model backbones that are significantly smaller than Resnet-18 for most class incremental scenarios: a reduced AlexNet variant (s-CIFAR100), and TinyNet (TinyImagenet), whereas we adopt Resnet-18 only for the two most complex scenarios (5 datasets and Imagenet100). In case of task incremental we are using similar achitecture: a 2-layer
neural network with fully connected layers (p-MNIST). Nevertheless, we are able to achieve a better performance in terms of accuracy in all these scenarios.
A comparison of the experimental results show that our method outperforms NPCL in terms of Accuracy in different scenarios: S-CIFAR-100 (77.27 vs. 71.34), TinyImagenet (80.10 vs. 60.18), and p-MNIST (97.14 vs. 95.97).

As for QDI \cite{madaan2023heterogeneous}, model backbones adopted in the paper are comparable to ours. 
Comparing the results shows that our method outperforms QDI on TinyImagenet (80.10 vs. 66.79) and QDI outperforms our method on s-CIFAR100 (77.27 vs. 88.30). 
However, one important consideration is that the setup used for TinyImagenet is 10 tasks with 20 classes, whereas our setting is 40 tasks with 5 classes. 
%
At the same time, the setup used in QDI for s-CIFAR100 presents 20 tasks with 5 classes, whereas in our work we consider 10 tasks with 10 classes. 
Moreover, considering accuracy in isolation may lead to a reductive analysis. A potential pitfall in the QDI paper is that there is no discussion (or empirical analysis) about model capacity (NPCL runs the models with similar size but there is no additional step of capacity reduction). This makes it difficult to gauge the trade-off between accuracy and memory impact of the method, which is one of our distinctive goals. 

It is also worth noting that NPCL \cite{jha2024npcl} and QDI \cite{madaan2023heterogeneous} papers did not provide experiments with the two most complex scenarios (5 datasets and Imagenet100) considered in our study, where our method achieves a remarkable performance.

\subsection{Ablation analysis}
In regards to model capacity, our findings reveal that our pruning step, which involves weight sharing, is directly responsible for reducing memory footprint. As a result, comparing our approach to Ada-QPacknet and WSN leads to a significant improvement in capacity (e.g. from $81.25\%$ and $77.73\%$ to $22.65\%$ for p-MNIST -- other examples are presented in Table 1). 
We argue that in case of Ada-QPacknet the quantization step 
contributes to model capacity to a lesser degree than adaptive pruning, since the bit-width achieved in our experiments is comparable to Ada-QPacknet \cite{pietron2023ada}. The quantization step contributes significantly when we try to compare TSN with WSN (WSN has no reduction stage for the bit-width of the parameters and activations).
Fine-tuning just minimally contributes to model capacity.

As for model accuracy, we argue that  performance gains are obtained thanks to our pruning approach with dynamic masks updated during the training process. This approach provides a significant improvement over other methods where masks are statically initialized, such as Ada-QPacknet (see Table 2).

\subsection{Open challenges}
As for potential limitations of the proposed approach, we envision that if the divergence between tasks is too high (low inter-task similarity) adaptive pruning may struggle to identify suitable shared weights, leading to memory saturation to accommodate all tasks. This possible limitation can be also identified in other continual learning methods such as WSN \cite{kang2022} and Ada-QPacknet \cite{pietron2023ada}. 
Similar considerations can be drawn for very long scenarios (e.g. characterized by hundreds of tasks). 
In this particular case, adaptive pruning could be less effective, and quantization is expected to be the main driver for memory reduction.
In general, future research is required to assess the robustness of continual learning methods in very complex scenarios.

\begin{table}[h!]
\small
\centering
\caption{Memory capacity results for codebooks, masks and weights (including pruning and quantized values). 
}
\begin{tabular}{|c|c|c|c|} 
\hline
\textbf{Scenarios}   & \textbf{weights} & \textbf{codebooks} &  \textbf{masks} \\ 
\hline
\textbf{p-MNIST}    &     9.8\%                    &   < 0.1\%   &        12.5\%         \\ 
\textbf{CIFAR100}        &   5.49 \%                      &    < 0.1\%      &                  12.5\%                  \\ 
\textbf{5 datasets}        &    14.57\%                      &    < 0.1\%      &    10\%                               
\\
\textbf{TinyImagenet}      &  12.5\%                     &     < 0.1\%      &  20\%             \\ 
\textbf{Imagenet100}      &   12.5\%                  &    < 0.1\%       &     12,5\%        \\ 


\hline
\end{tabular}
\label{table:capacity_codebooks_masks_weights}
\end{table}

\begin{table*}[h!]
\small
\centering
\caption{Sparsity and bit widths for different benchmarks.}
\begin{tabular}{|c|c|c|c|} 
\hline
\textbf{Scenarios}   & \textbf{sparsity before fine tuning} & \textbf{sparsity after fine tuning} &  \textbf{bit-width} \\ 
\hline
\textbf{p-MNIST}    &       37.58                 &   41.2   &      4 bits         \\ 
\textbf{CIFAR100}        &     7.22                  &    11.68    &         4 bits                          \\ 
\textbf{5 datasets}        &    1.02                      &     21.02     &        4 bits                          
\\
\textbf{TinyImagenet}      &     2.7                  &     14.31      &       5 bits        \\ 
\textbf{Imagenet100}      &          8.5           &     10.1     &      5 bits      \\ 


\hline
\end{tabular}
\label{table:sparsity}
\end{table*}

\begin{table*}[h!]
\small
\centering
\caption{Accuracy results for different weight bit width.}
\begin{tabular}{|c|c|c|c|} 
\hline
\textbf{Scenarios}   & \textbf{5 bits} & \textbf{4 bits} &  \textbf{3 bits} \\ 
\hline
\textbf{p-MNIST}    &        96.85                 &        96.63          &       96.06          \\ 
\textbf{CIFAR100}        &   74.82              &   75.14     &   75.07                                          \\ 
\textbf{TinyImagenet}      &       77.19               &     77.16      &     68.01          \\ 
\textbf{5 Datasets}      &        91.71            &       91.80   &       90.1     \\ 
\hline
\end{tabular}
\label{tab:appendix-accuracy-bit-width}
\end{table*}

\begin{table*}[h!]
\small
\centering
\caption{Average execution times on single GPU for different benchmarks (in seconds).}
\begin{tabular}{|c|c|c|} 
\hline
\textbf{Scenarios}   & \textbf{training time [s]}  &  \textbf{fine tuning time [s]} \\ 
\hline
\textbf{p-MNIST}    &   1 244                 &              23.92                   \\ 
\textbf{CIFAR100}        &     662   &                     12.6                                \\ 
\textbf{TinyImagenet}      &    6752              &         122.76                 \\ 
\textbf{Imagenet100}      &         5 094            &        84.9                \\ 
\hline
\end{tabular}
\label{tab:time}
\end{table*}

\begin{table*}[h!]
\small
\centering
\caption{Comparison of capacity with and without post-pruning (post-pruning run with 100 iterations), and with replay memory.}
\begin{tabular}{|c|c|c|c|} 
\hline
\textbf{Scenarios}   & \textbf{without post-pruning} & \textbf{with post-pruning} &  \textbf{with replay memory} \\ 
\hline
\textbf{p-MNIST}    &         23.41\%             &    22.65\%     &   37.5\%* \\ 
\textbf{CIFAR100}        &       
18.75\%&     17.62\%  & 41.92\% \\ 
\textbf{TinyImagenet}      &   36.33\%             &     32.15\%  &  93.6\%\\ 
\textbf{5 Datasets}      &    30.62\%  &  24.68\%   &   40.08\%\\ 
\hline
\end{tabular}
\label{tab:appendix-capacity-post-pruning-replay}
\end{table*}


\begin{table*}[h!]
\small
\centering
\caption{Number of FLOPs in case of floating point and with reduced bit-width of both weights and activations.}
\begin{tabular}{|c|c|c|c|c|c|c|c|} 
\hline
\textbf{Scenarios}   & \textbf{4 bit + p} &\textbf{8 bit + p} & \textbf{16 bit + p} & \textbf{4 bit} & \textbf{8 bit} & \textbf{16 bit} & \textbf{baseline floating-point} \\ 
\hline
\textbf{p-MNIST}   & 0.97M &    3.89M     &   15.6M  &   1.62M   &   6.5M       &       26M      &    61.7M\\ 
\textbf{CIFAR100}   & 214.37M &  857.5M  &  3 430M &  243.6M &    974.4M         &    3 897M     & 8 770M \\ 
\textbf{TinyImagenet}  &  112.72M  & 450.9M  &  1 803M &  132.62M &  530.5M              &  2 122M     & 4 775M\\ 
\hline
\end{tabular}
\label{tab:appendix-flops}
\end{table*}


\begin{table*}[h!]
\centering
\caption{Accuracy with reduced bit-width of activations and quantized centroids.}
\begin{tabular}{|c|c|c|c|} 
\hline
\textbf{Scenarios}   & \textbf{4 bit} & \textbf{8 bit} & \textbf{16 bit} \\ 
\hline
\textbf{p-MNIST}    
& 84.08  &   97.06     &    97.13     \\ 
\textbf{CIFAR100}       
 & 70.86 &   75.21   &   75.46 \\ 
\textbf{TinyImagenet}      
&  75.84  &  79.66   &  79.70\\ 
\textbf{5 Datasets}     
&  85.2 &  91.71   &  91.78 \\ 
\hline
\end{tabular}
\label{tab:flops_acc}
\end{table*}

\begin{table*}[h!]
\small
\centering
\caption{Forward and Backward Transfer for TSN}
\begin{tabular}{|c|c|c|c|} 
\hline
\textbf{Scenarios}   & \textbf{Backward Transfer} & \textbf{Forward Transfer}  \\ 
\hline
\textbf{p-MNIST}    & 0.0                     &  0.104                       \\ 
\textbf{CIFAR100}        &  0.0               &      0.010                                           \\ 
\textbf{TinyImagenet}      &  0.0                 & 0.004               \\ 
\textbf{5 Datasets}      &  0.0                 &   -           \\ 
\textbf{Imagenet100}      &  0.0                &      -        \\ 
\hline
\end{tabular}
\label{tab:appendix-forward-backward}
\end{table*}

\begin{figure*}[ht]
   \centering
    \includegraphics[width=0.75\textwidth]{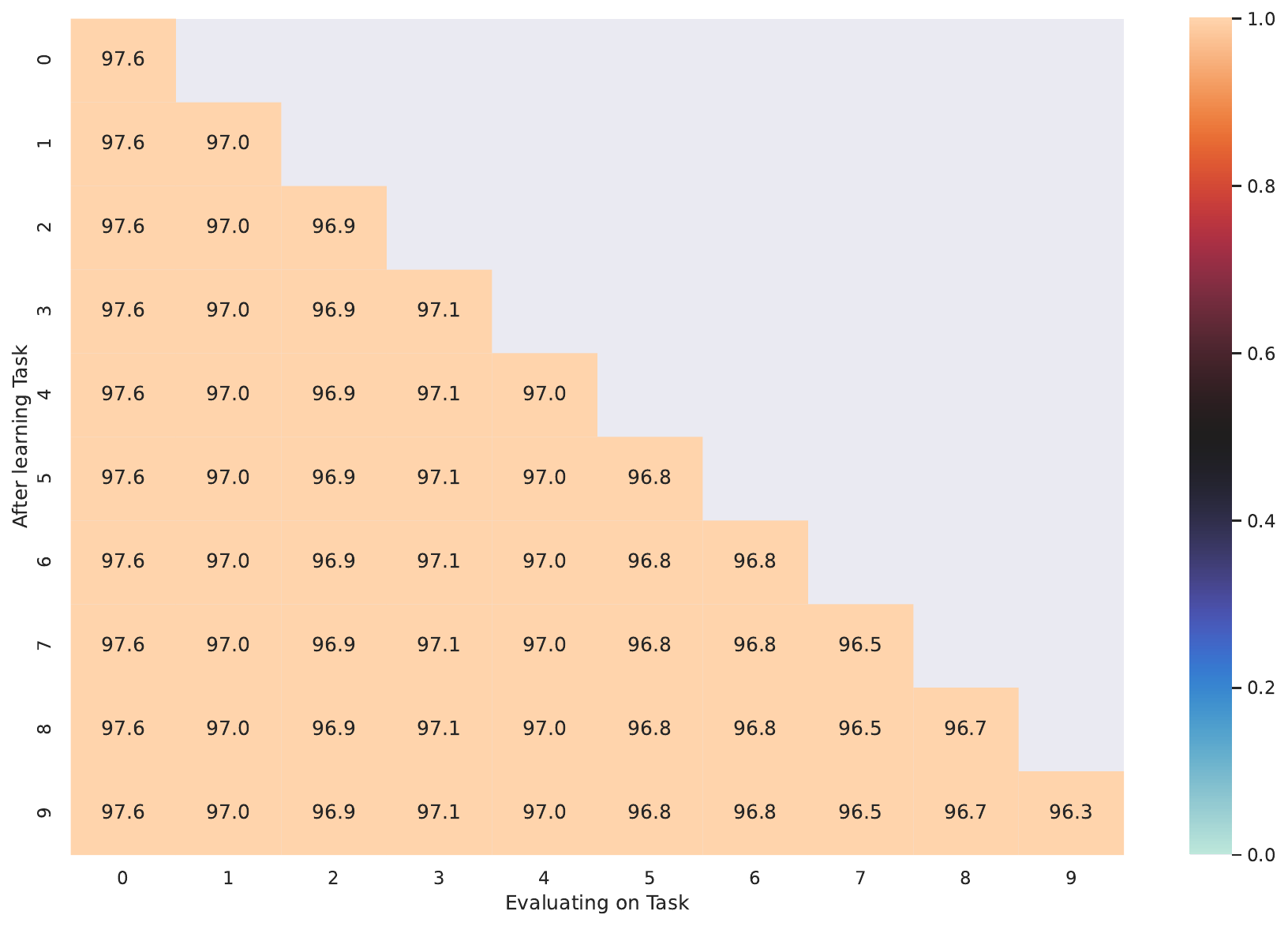}
  \caption{Accuracy matrix for p-MNIST (TSN-wr)}
    \label{fig:heatmap-pmnist}
\end{figure*}

\begin{figure*}[ht]
   \centering
    \includegraphics[width=0.75\textwidth]{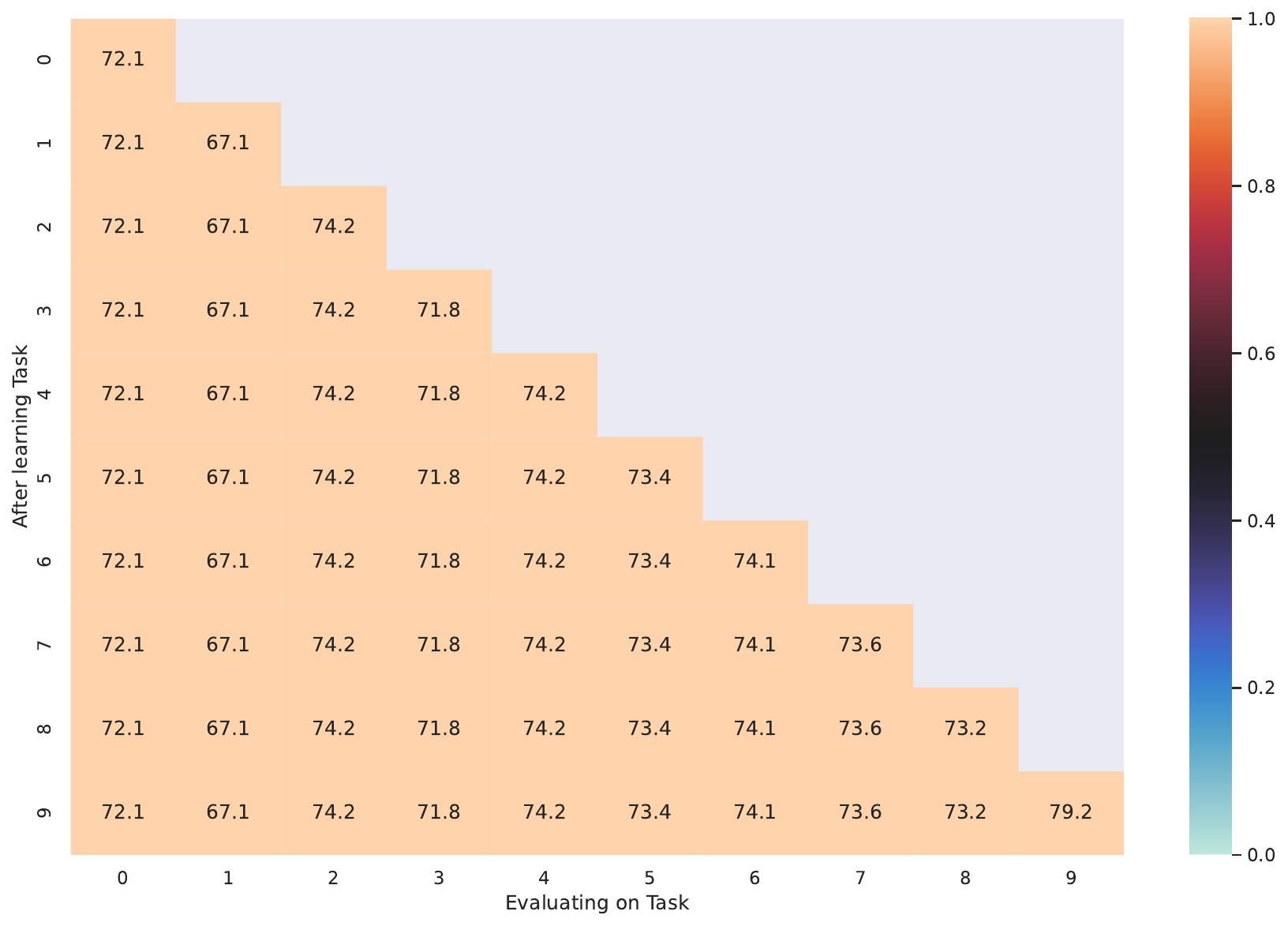}
  \caption{Accuracy matrix for CIFAR100 (TSN-wr)}
    \label{fig:heatmap-cifar}
\end{figure*}

\begin{figure*}[ht]
   \centering
    \includegraphics[width=0.75\textwidth]{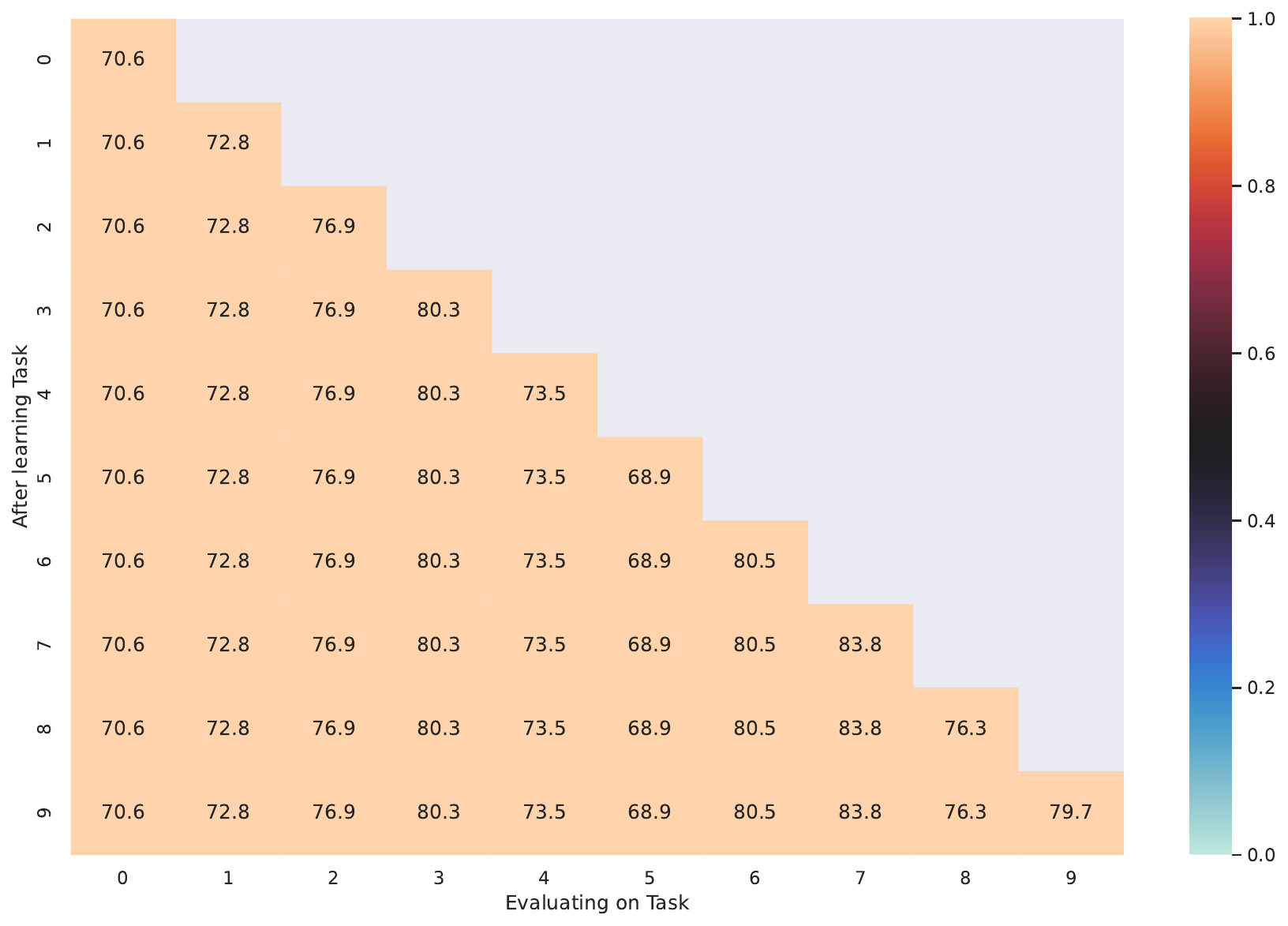}
  \caption{Accuracy matrix for Imagenet-100 (TSN-wr)}
    \label{fig:heatmap-imagenet-100}
\end{figure*}


\begin{figure*}[ht]
   \centering
\includegraphics[width=0.75\textwidth]{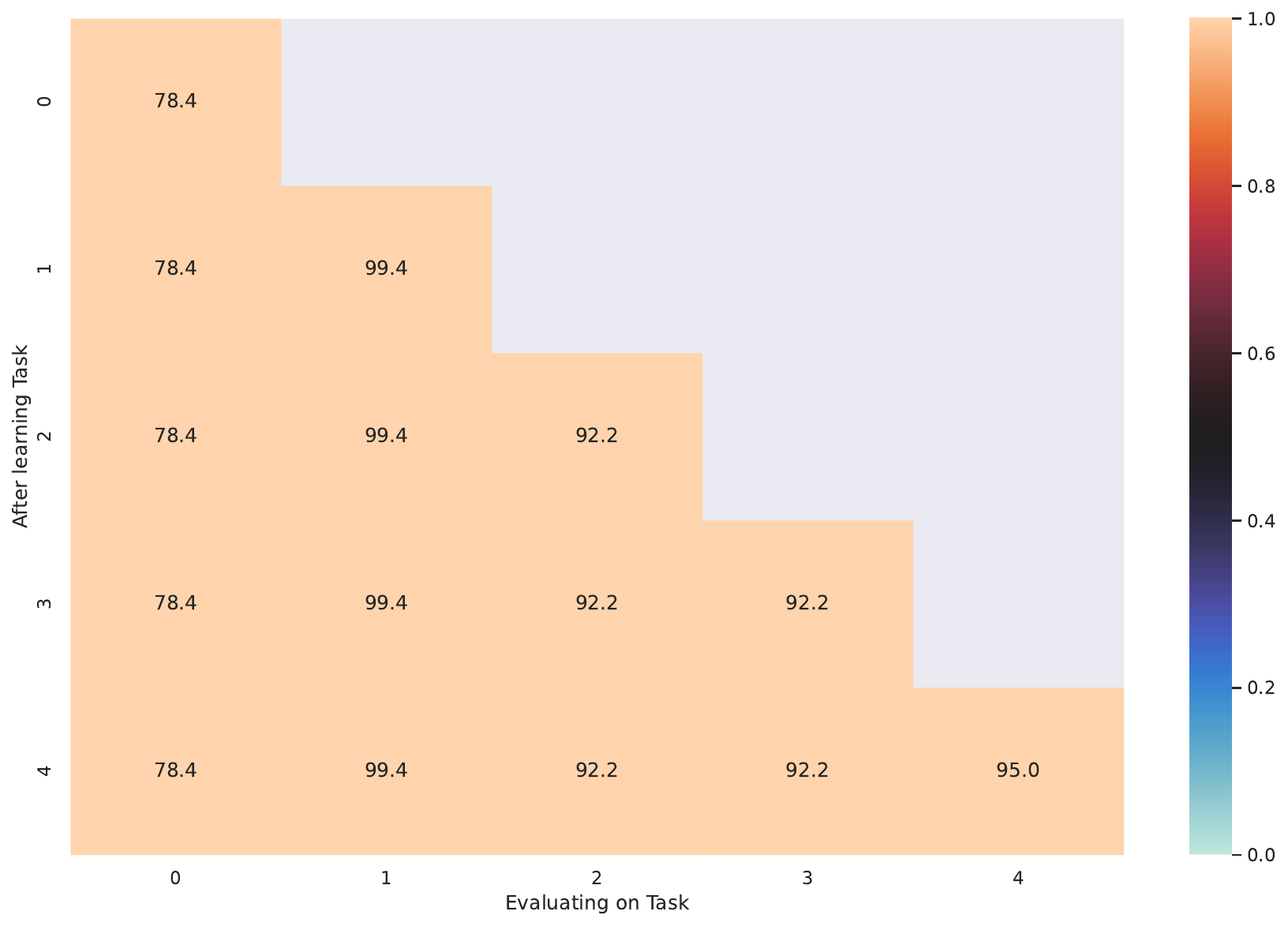}
  \caption{Accuracy matrix for 5 datasets (TSN-wr)}
    \label{fig:heatmap-5datasets}
\end{figure*}

\begin{figure*}[ht]
   \centering
    \includegraphics[width=\textwidth]{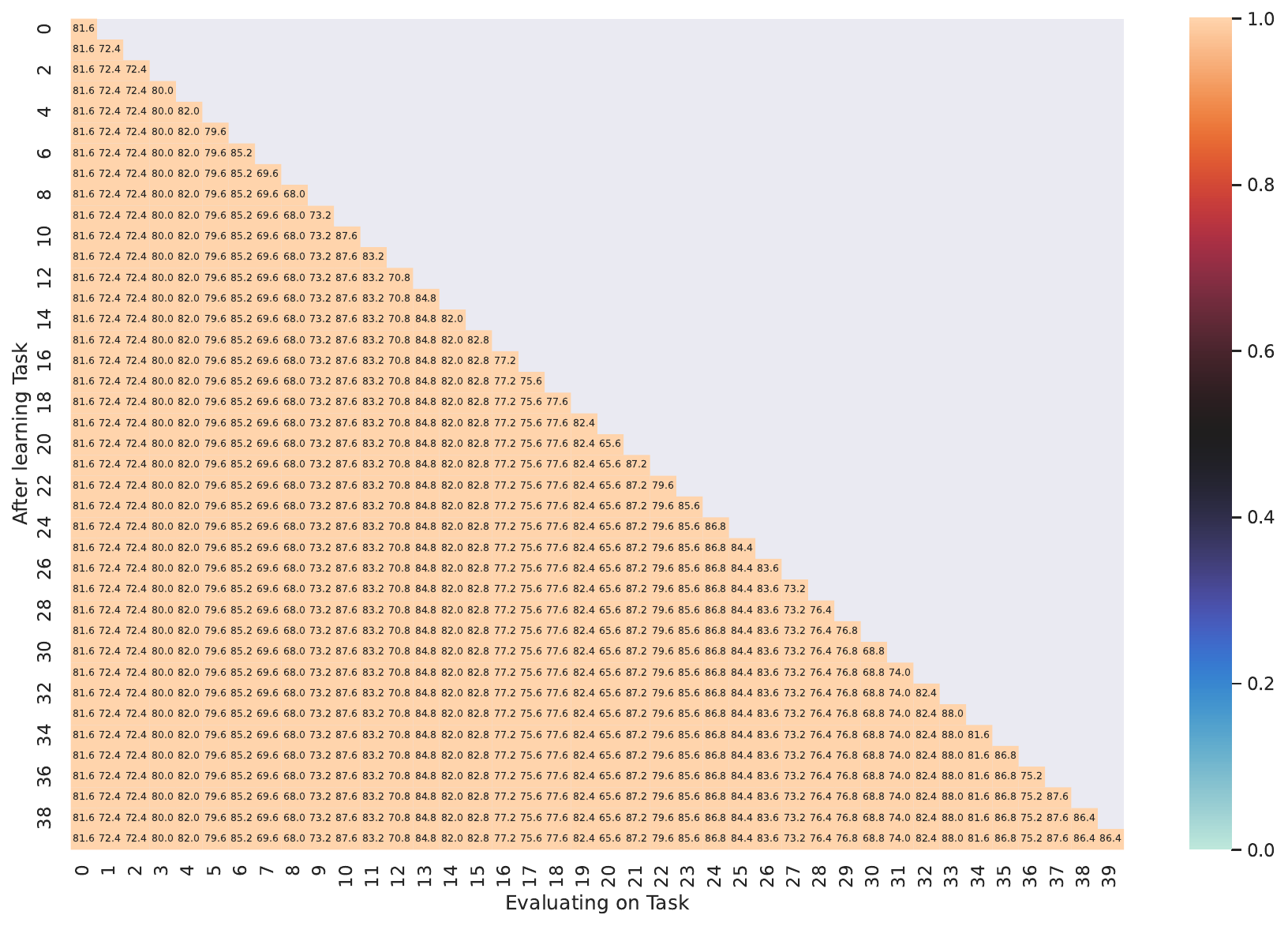}
  \caption{Accuracy matrix for TinyImagenet (TSN-wr)}
    \label{fig:heatmap-imagenet}
\end{figure*}

\end{document}